\documentclass[a4paper]{cas-dc}
\usepackage{amsthm}
\usepackage{ulem}
\usepackage{amssymb}
\let\emptyset\varnothing

\usepackage[numbers]{natbib}
\def\tsc#1{\csdef{#1}{\textsc{\lowercase{#1}}\xspace}}
\tsc{WGM}
\tsc{QE}
\tsc{EP}
\tsc{PMS}
\tsc{BEC}
\tsc{DE}

\begin{document}
\let\WriteBookmarks\relax
\def\floatpagepagefraction{1}
\def\textpagefraction{.001}
\shorttitle{DGHM}
\shortauthors{T.Lin, Y.Guo, et~al.}

\title [mode = title]{Decoupled Gradient Harmonized Detector for Partial Annotation: Application to Signet Ring Cell Detection}                      



\author[1]{Tiancheng Lin}[type=author,auid=000,bioid=1,]
\fnmark[1]
\ead{ltc19940819@sjtu.edu.cn}

\author[1]{Yuanfan Guo}[type=author,
                        auid=000,bioid=1,
                        orcid=0000-0002-5835-8545,
                        ]
\ead{gyfastas@sjtu.edu.cn}
\fnmark[1]
\author[1]{Canqian Yang}[type=author,
                        auid=000,bioid=1,
                        ]
\ead{charles.young@sjtu.edu.cn}
\author[1]{Jiancheng Yang}[type=author,
                        auid=000,bioid=1,
                        ]
\ead{jekyll4168@sjtu.edu.cn}
\author[1]{Yi Xu}[type=author,
                        auid=000,bioid=1,
                        ]
\cormark[1]
\ead{xuyi@sjtu.edu.cn}

\address[1]{Shanghai Jiao Tong University, School of Electronic Information and Electrical Engineering }









\cortext[cor1]{Corresponding author}
\fntext[fn1]{These authors are contributed equally to this work}


\begin{abstract}
Early diagnosis of signet ring cell carcinoma dramatically improves the survival rate of patients. Due to lack of public dataset and expert-level annotations, automatic detection on signet ring cell (SRC) has not been thoroughly investigated. In MICCAI DigestPath2019 challenge, apart from foreground (SRC region)-background (normal tissue area) class imbalance, SRCs are partially annotated due to costly medical image annotation, which introduces extra label noise. To address the issues simultaneously, we propose Decoupled Gradient Harmonizing Mechanism (DGHM) and embed it into classification loss, denoted as DGHM-C loss. Specifically, besides positive (SRCs) and negative (normal tissues) examples, we further decouple noisy examples from clean examples and harmonize the corresponding gradient distributions in classification respectively. Without whistles and bells, we achieved the 2nd place in the challenge. Ablation studies and controlled label missing rate experiments demonstrate that DGHM-C loss can bring substantial improvement in partially annotated object detection. 

\end{abstract}




\maketitle

\section{Introduction}
Signet ring cell carcinoma (SRCC) is a form of highly malignant adenocarcinoma. Nuclei push against cell membranes creating a typical signet ring cell (SRC), which has a large mucin vacuole filling the cytoplasm \cite{2000world}. SRCC tumors are mostly found in stomach and less frequently in breast, bladder and other organs \cite{organs}. Since the prognosis for patients with SRCC is extremely poor, early diagnosis and aggressive surgical intervention are of essential importance. However, manual analysis of digital pathology image is labor-intensive and time-consuming, which has become a bottleneck to diagnosis. Therefore, computer-aided technical for SRC detection, as an ancillary study, is promising and highly-demanded. 

MICCAI DigestPath2019 is the first challenge and first public dataset on SRC detection. Before this challenge, automatic algorithms on SRC detection have not been thoroughly investigated. However, perfect annotations are practically impossible due to various appearances of SRCs, as well as scattered/overcrowded regions. Thus, the dataset in the challenge is partially annotated and the task is partially annotated object detection (PAOD). As can be seen in Figure 1, in the abnormal pathology (AP) image, green and yellow bounding boxes denote annotated and unannotated SRCs respectively, while normal pathology (NP) image contains no SRCs at all. In conventional detection task \cite{fasterrcnn,focal}, there are only positive and negative examples.
However, when it comes to PAOD, some positive examples will be incorrectly labeled as negative, which introduces label noise. 
Directly training with noisy labels will cause adverse impact to the classification accuracy \cite{adversial1,adversial2}.

One solution, proposed by the challenge organizers, is a semi-supervised learning framework for label correction \cite{organizer}, which consists of three steps: initial fully-supervised training, self-training and co-training. Experiments demonstrate that not only is annotation quality improved, extra unlabeled images are also better explored. 
Instead of pseudo label generation, we tackle PAOD in a loss correction paradigm. The basic idea is to decouple noisy examples from clean examples, disentangling the partial annotation into two parts of noisy-supervised learning and full-supervised learning. However, there are two intrinsic issues: noisy example overfitting and hard example under-learning. Several studies address these two problems respectively. On one hand, Huber loss \cite{huber} and GHM-C loss \cite{GHM} consider hard examples mostly as outliers and reduce their loss contribution. Thus, noisy example overfitting can be prevented, which is suitable for noisy-supervised learning. But they also sacrifice the ability of models to learn hard examples well. On the other hand, online hard example mining \cite{OHEM} and focal loss \cite{focal} put more emphasizes on hard examples. Hence, hard example under-learning can be addressed, which is appropriate for full-supervised learning. Unfortunately, these methods further aggravate the overfitting of noisy examples. 

Inspired by GHM \cite{GHM}, we propose Decoupled Gradient Harmonising Mechanism (DGHM) for PAOD, addressing the problems of hard-example under-learning and noisy-example overfitting at the same time.
We apply DGHM by decoupling noisy examples from clean examples, harmonizing their gradient distributions separately and managing their outliers differently. 
For the classification branch of detection, DGHM is embedded into cross entropy (CE) loss, forming DGHM-C loss. Particularly, hard examples in noisy part will be explicitly down-weighted to prevent noisy example overfitting, while the opposite operation is performed on clean part to address hard example under-learning. 
For the regression branch, since regression loss is only calculated over positive examples, we take smooth-$L_1$ loss for simplicity. 
Experiments show that with the help of DGHM, the performance of PAOD is substantially improved.
The main contributions of our work are summarized as follows:
\begin{enumerate}
    \item We decouple noisy examples from clean examples and disentangle partial annotation into noisy-supervised learning and full-supervised learning.
    \item We propose the DGHM and DGHM-C loss for PAOD, making the model robust to noisy examples and focus on hard case of clean examples simultaneously.
    \item We conduct comprehensive experiments to demonstrate the effectiveness of DGHM-C loss, which shows an improvement over other loss functions.
\end{enumerate}

\begin{figure}
\setlength{\abovecaptionskip}{-0.cm}
\setlength{\belowcaptionskip}{-0.cm}
\begin{center}
\includegraphics[width=0.5\textwidth]{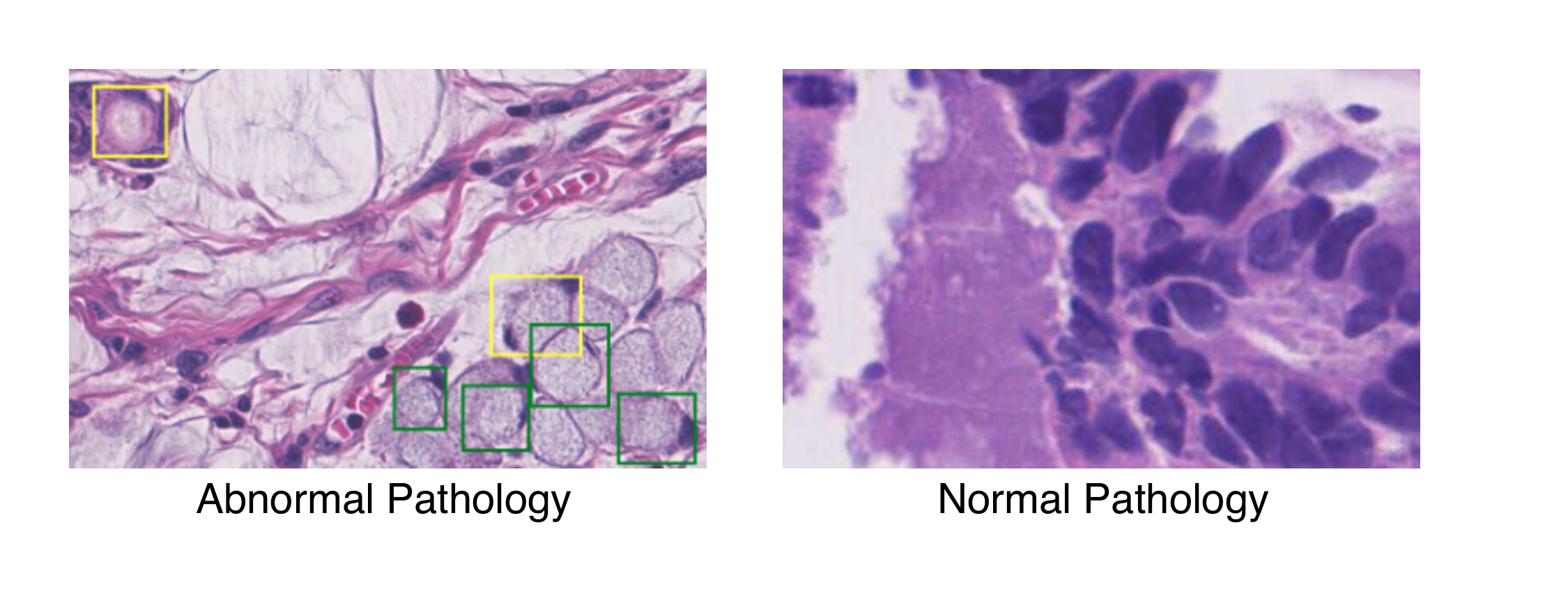}
\caption{
Examples of abnormal and normal pathology image provided in our dataset. Signet ring cell only exists in AP image. The green bounding boxes are provided annotation while the yellow one might be signet ring cell but not annotated.
}

\label{Fig1.}

\end{center}
\end{figure}

\section{Related Work}

\subsection{Object Detection}
Object detection is one of the most fundamental and important task in the field of computer vision, which involves classifying and locating the objects. Deep neural networks have made great progress in object detection \cite{fasterrcnn,fpn,rcnn,fastrcnn,R-FCN}.
Generally, detectors are divided into one-stage and two-stage according to whether to skip the region proposal stage as well as anchor-based and anchor-free depending on whether to use predefined anchors which serve as hypotheses over possible locations of ground-truth bounding boxes. Here, we classify detectors into full-supervised, weakly supervised and noisy-supervised with regard to the quality of labels.
\paragraph{Full-supervised Detection}
In fully-supervised object detection (FSOD), focal loss \cite{focal} and GHM-C loss \cite{GHM} are proposed to tackle the class imbalance problem in  positive/ negative and hard/easy examples. Guided Anchoring \cite{GA} leverages semantic features to generate high-quality proposals and boosts the detection performance. Libra RCNN \cite{Libra} designs an overall balanced learning, consisting in sample level, feature level and object level, which can be generalized to both one-stage and two-stage detectors. 
However, these methods highly depend on fine-grained instance-level bounding boxes.
\paragraph{Weakly Supervised Detection}
Weakly supervised object detection (WSOD) requires only image level annotation. 
WSDNN \cite{WSDNN} designs a specific framework with two streams of recognition and detection, and the final proposals are jointly supervised by the two streams. Based on WSDNN, contextual region \cite{Contextlocnet} and multi-stage classifier refinement \cite{multiRefine} are introduced for further improvement. 
Recently, a min-entropy latent model \cite{wan2018min} is proposed to discover latent objects and minimize the localization randomness for WSOD with recurrent learning algorithm.
Moreover, Multiple Instance Learning (MIL), as a common method for WSOD, is recently further improved in C-MIL \cite{C-MIL} by alleviating the non-convexity problem.
Some works also use additional annotations or data to improve
the performance, e.g. domain adaption \cite{LSDA,ProgressiveDA}, semantic knowledge transfer \cite{SKT} and so on.

\paragraph{Noisy-supervised Detection}

Different from WSOD, in noisy-supervised detection, models are trained with partial bounding-box annotations, which is called PAOD in this paper. In \cite{RobustNucleus}, a self-attention mechanism is adopted to handle the partial annotation problem in Nuclei detection task. 
A hybrid supervised learning framework \cite{MissingLabel} is proposed to solve the missing label problem by generating pseudo labels during training. 
Moreover, soft sampling strategy \cite{Soft_sample} is established to alleviate the effect of overcrowded false negatives, while giving no guarantee to scattered false negatives.
{Co-occurrence loss is introduced in \cite{PFDet} to solve the problem of sparsely verified classes, where hierarchical and spatial relationships between multiple classes are exploited to alleviate noisy example overfitting. It imposes the label consistency constraint on the subclass and its parent class, where the performance depends on the high-quality labels of the parent classes. In the DigestPath2019 challenge, however, noisy labels only exist in the class of SRC due to partial annotation in the AP images. Meanwhile, clean labels are available for the annotated SRCs and the NP images. Accordingly, this paper contributes to both reducing the overfitting problem for the noisy labels of SRC in the AP images and the under-learning problem of hard examples with the clean labels.}



\subsection{Learning with Noisy Examples}
Learning with noisy examples is a common situation especially in medical imaging community where manual annotation could be expensive and needs expert knowledge \cite{unreliableannotaion}. In the existing literature, the solutions can be roughly classified into two types: label correction and robust loss function.

\paragraph{Label Correction}
Some previous works tackle the noisy label problem by removing or correcting the noisy labels during training. 
In \cite{mentornet,metacleaner,minimumsupervision,CleanNet}, auxiliary networks are introduced to correct the missing labels. For example, in \cite{CleanNet}, a feature encoder that learns the class representative embedding vectors (called "class-prototype") is utilized to decide whether the label is correct or not by comparing with the prototype. MentorNet \cite{mentornet} pre-trains an extra network for clean label selection. In \cite{metacleaner}, an extra module called MetaCleaner is used to assign scores for samples labeled as same category and hallucinate a clean representation based on the scores.
Recently, specific training strategies such as self-training \cite{organizer,jointoptimization,reed2014bootstrapping} and curriculum learning \cite{curriculumnet,multilabelcurriculum,LwithNoise} are proposed for learning under presence of noise. In \cite{organizer}, co-training and self-training are both used to iteratively generate pseudo labels, which shows great performance in SRC detection task.
Knowledge distillation is first proposed in \cite{knowledgedistillation} and explored in \cite{LearningfromNoisyDist} to handle the noisy label problem, which utilizes knowledge learned from an extra clean dataset to generate soft labels for a larger noisy dataset. However, all these methods require either auxiliary network architectures, complex training procedure or additional supervision, which makes it hard to be deployed in real-world applications.

\paragraph{Robust Loss Function}
Robust loss function has been widely employed in learning with noisy examples. Backward \cite{ForwardBackword} and Forward \cite{ForwardBackword} modify the loss function depending on the noise transition matrix. Nonetheless, the accurate estimation of transition matrix is not promised. Various regularizers, like label smoothing \cite{LSR} and confidence penalty \cite{penalty}, are introduced to prevent noisy example overfitting at the expense of hard example under-learning. 
Moreover, Symmetric Cross Entropy (SCE) \cite{SCE} combines CE with a noise tolerant term, so that both hard example learning and noisy example robustness are improved. However, directly using robust loss in SRC detection is unreasonable, because partial annotation in AP leads to the coexistence of noisy examples and clean examples.

\paragraph{}
Therefore, we establish DGHM and DGHM-C loss for PAOD. Without any auxiliary, the problems of hard example under-learning and noisy example overfitting can be tackled simultaneously by decoupling noisy examples from clean ones. The proposed DGHM-C loss can be directly applied to SRC detection in MICCAI DigestPath2019 challenge.


\section{Method}

\begin{figure*}
\setlength{\abovecaptionskip}{-0.cm}
\setlength{\belowcaptionskip}{-0.cm}
\begin{center}
\includegraphics[width=0.95\textwidth]{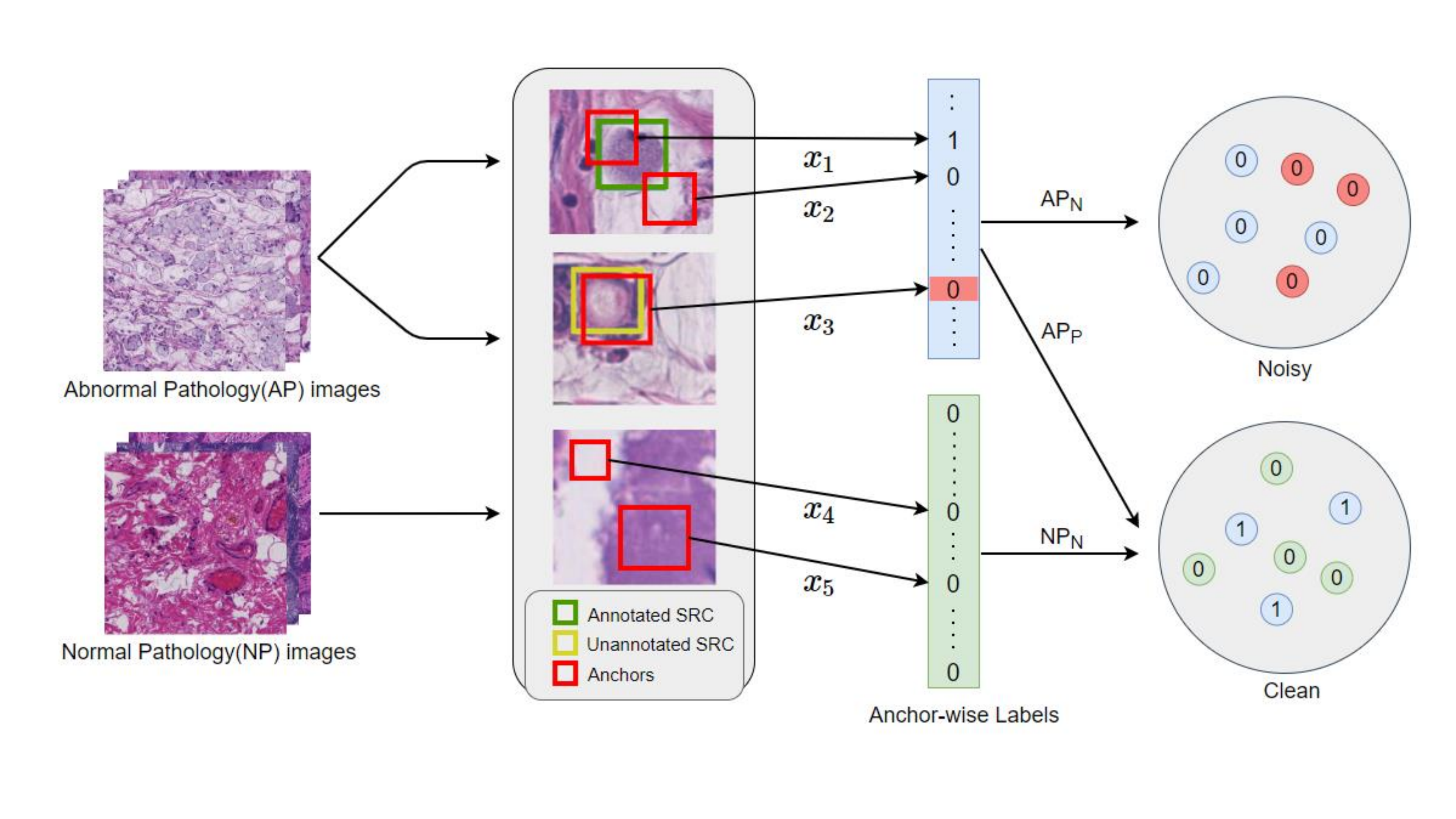}
\caption{
Illustration of noisy labels introduced in partially annotated SRC detection. Since anchors (such as $x_3$) matching unannotated bounding boxes (yellow) are { falsely} labeled as negative { examples}, negative examples in AP images ($\textit{\text{AP}}_N$) are noisy, where some are actually positive. Since annotated bounding boxes (green) are indeed SRCs and it is certain that no SRCs in NP images, positive examples in AP images ($\textit{\text{AP}}_P$) and negative examples in NP images($\textit{\text{NP}}_N$) are clean examples.
}

\label{Fig2.}

\end{center}
\end{figure*}

\subsection{Problem Description}

In MICCAI DigestPath2019, automatic algorithms are demanded for SRC detection with partial annotation. 
{Using assignment rule from \cite{focal}, label $p^*_i \in \{0,1\}$ is assigned to the candidate anchor with index i, where '0' denotes negative examples while '1' represents positive examples.} As shown in Figure 2, the green boxes and yellow boxes represent annotated and unannotated SRCs, respectively. The red boxes indicate several candidate anchors, denoted as $x_i$. Suppose anchor $x_1$ has an Intersection-over-Union(IoU) overlap with one green box over 0.5, then it is labeled as '1'. In contrast, anchor $x_2$ is labeled as '0', since their IoU overlap is less than 0.5. The anchors in NP images, i.e. $x_4$ and $x_5$, will be all labeled as 0, since there is no SRCs existed. However, though anchor $x_3$ has an IoU with one yellow box over 0.5, it will be mislabeled as '0'. { Actually, yellow box is missed due to partial annotation of SRCs in AP images. Thus it can be seen, for PAOD, labeling anchors only based on IoU is inaccurate for classification. Specifically, label noise will be only introduced to the negative examples of AP images. Both the positive examples of AP images and negative examples spreading over NP images are clean examples with experts’ manual annotation.}

Fortunately, the noisy anchors can be easily located by introducing an extra attribute into the dataset. Generally, more attributes contribute to better describing the dataset, which may further improve the performance of downstream tasks. Back to the partial annotation problem, we assign image-wise label to each anchor as an extra attribute. By combining the attribute with anchor-wise label, noisy examples can be decoupled from clean examples, as can be seen in Figure 2. Specifically, negative examples from AP images ($\textit{\text{AP}}_N$) consists of the noisy class, because both correctly labeled anchors such as $x_2$ and incorrectly labeled anchors such as $x_3$ exist in these examples, since only a small portion of SRCs are annotated. Meanwhile, positive examples from AP images ($\textit{\text{AP}}_P$) 
and negative ones from NP images ({ $\textit{\text{NP}}_N$}) consists of the clean class. Since all annotated cells are indeed SRCs and NP images contain no SRCs at all.

With the attribute of image-wise label $a_i$, the training data space can be defined as $S=\{(x_i, p^*_i, a_i)\}_{i=1}^N$, where $N$ is the total number of anchors {  and $p^*_i$ is the given label for example with index i}. Noisy and clean data space can also be obtained by $S_{n} = \{(x_i,p_i^*,a_i)\text{ }|\text{ }p_i^*=0 \text{ and } a_i=1\}_{i=1}^{N_{n}}$ and $S_{c} = \{(x_i,p_i^*,a_i)\text{ }|\text{ }p_i^*=1 \text{ or } a_i=0\}_{i=1}^{N_{c}}$, subject to $N_{n}+N_{c} = N$, $S_{n} \cap S_{c} = \emptyset$ and $S_n \cup S_c=S$, where $N_n$ and $N_c$ are the number of noisy and clean anchors. We assume that there is a corresponding ideal label $\hat{p^*}$, thus noisy and clean examples could be described as:
\begin{equation}
    P(\hat{p_i^*} \neq p_i^*) = \left\{
    \begin{array}{ll}
    \eta, & \text{if } x_i \in S_n \\
    0, & \text{if } x_i \in S_c.
    \end{array}
    \right.
\end{equation}
We denote $P(\hat{p_i^*} \neq p_i^*)$ as the probability that the given label is incorrect where $\eta$ is label corruption ratio, which is nearly proportional to the number of unannotated cells. For clean anchors the label corruption ratio is 0.

\subsection{Classification Loss}
Object detection includes two sub tasks: classification and bounding boxes regression\cite{fasterrcnn,fastrcnn,R-FCN}. To alleviate the problem mentioned above, our efforts are mainly focused on the classification sub task. As a matter of fact, our method is established based on focal loss and GHM-C loss, which are recently introduced to address the class imbalance from different aspects. Both of them are based on CE loss, which is defined as
\begin{equation}
    L_{CE}(p,p^*) = \left\{
    \begin{array}{ll}
    -\log(p), & { p^*=1} \\
    -\log(1-p), & { p^*=0.}
    \end{array}
    \right.
\end{equation}
{In eq.(2), $p^* \in\{0,1\}$ denotes the given ground-truth label and $p\in[0,1]$ denotes the probability of the class with label $p^*=1$ according to the model estimation.}
The gradient norm \textit{g} is defined as
\begin{equation}
g = |p-p^*| = \left\{
    \begin{array}{ll}
    1-p,          & p^*=1 \\
    p,         & p^*=0.
    \end{array}
    \right.
\end{equation}
Correspondingly, focal loss can be formulated as
{
\begin{equation}
    L_{focal} = \frac{1}{N} \sum_{i=1}^{N}-\alpha_t g_i^\gamma L_{CE}(p_i,p_i^*) \text{.}
\end{equation}
Focal loss down-weights the contribution of easy examples by adding a weighting factor $\alpha_t$ and a modulating factor $g^\gamma$, where $\gamma \geq 0$ is a tunable focusing factor. For notation convenience, $\alpha_t$ can be simplified analogously to \textit{g} and assigned with $\alpha\in [0,1]$ for foreground and $1-\alpha$ for background.}
Besides easy examples, GHM-C loss further down-weights the contribution of outliers regarding to gradient density (GD) \cite{GHM}. GD denotes the number of examples in specific regions. GD function and the weighted parameters $\beta$ are defined as

\begin{equation}
    GD(g_i) = 
    \frac{1}{{l_\epsilon(g_i)}}\sum_{k=1}^N \delta_ \epsilon(g_k,g_i), \\
\end{equation}
\begin{equation}
\beta_i= \frac{N}{GD(g_i)},
\end{equation}
where $g_k$, $g_i$ denote the gradient norm for the example with index k and i respectively and $\delta_\epsilon(g_k,g_i)$ is a function indicating 
whether $g_k$ is located in the region centered at $g_i$ with the valid length  $l_\epsilon(g_i) = max(1,g_i+\frac{1}{2}\epsilon) - min(0,g_i-\frac{1}{2}\epsilon)$. Thus, GHM-C loss function is

\begin{equation}
\begin{aligned} L_{GHM-C} &=\frac{1}{N} \sum_{i=1}^{N} \beta_{i} L_{C E}\left(p_{i}, p_{i}^{*}\right).
\end{aligned}
\end{equation}

{Recently, Symmetric Cross Entropy(SCE) loss is introduced in \cite{SCE} to handle the hard example under-learning and the noisy example overfitting problem in classification. SCE loss is defined as the combination of CE and reversed CE.
\begin{equation}
    L_{SCE}(p,p^*) = \alpha_{SCE} L_{CE}(p,p^*) + \beta_{SCE} L_{CE}(p^*,p),
\end{equation}
where $\alpha_{SCE}$ and $\beta_{SCE}$ are weighting factors to balance between CE and the noise-robust term.}

\subsection{Motivation}
{
Intuitively, we can train a model with only clean examples. However, experimental results suffer from high false positive rate. That is mainly because model is not able to learn how to discriminate background (normal tissue areas) from foreground (SRC) in AP images without supervision on $\textit{\text{AP}}_N$. In fact, the normal tissue areas in AP images present a much different appearance from those in NP images. Consequently, the diversity of the background will not be fully learned in the training using only a part of dataset.}
However, if training with both clean and noisy examples, a dilemma arises when applying a unified function such as CE loss, focal loss or GHM-C loss. To illustrate the dilemma, we take a converged model trained with CE loss as an example.
Different from \cite{GHM}, we figure out the gradient norm distribution of the model in Figure 3, where the noisy and clean examples are counted in two independent histograms. Outliers (examples with large gradient norm) exist in both histograms yet with different meanings. In the histogram of clean examples, outliers are the hard examples. In contrast, the outliers in the histogram of noisy examples are very likely to be mislabeled examples. Moreover, outliers in clean examples are relatively more than that in noisy examples, which indicates that the model under learns the hard examples but overfits the noisy examples.  
Therefore, we propose DGHM to conquer noisy and clean examples separately, providing two different operations on the outliers corresponding to two kinds of examples.
  
\begin{figure}
\setlength{\abovecaptionskip}{0.cm}
\setlength{\belowcaptionskip}{-5.cm}
\begin{center}
\includegraphics[width=0.5\textwidth]{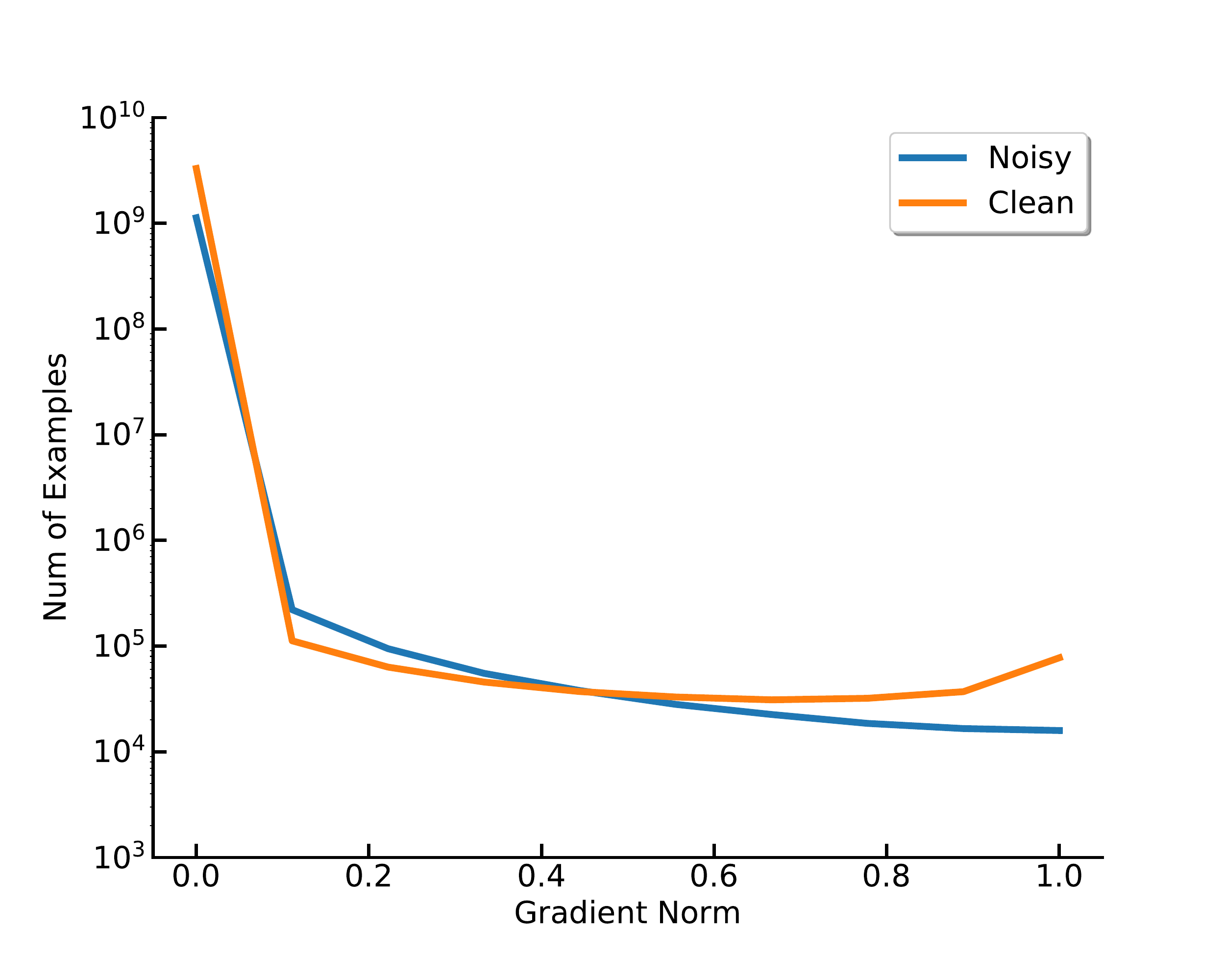}
\caption{An illustration of the gradient norm distribution of a converged model trained with Cross Entropy loss. The y-axis is in log scale. }
\label{Fig3.}

\end{center}
\end{figure}






\subsection{DGHM}
We propose DGHM for different outlier operations in PAOD. 
To be specific, we decouple noisy examples from clean ones and calculate their gradient norm distribution respectively, as shown in Figure 3. Then, the GD function is reformulated as


\begin{equation}
    GD(g_i) = \left\{
    \begin{array}{ll}
    \frac{1}{{l_\epsilon(g)}}(\sum_{k=1}^{N_{c}} \delta_ \epsilon(g_k,g_i)), & { x_i \in S_c} \\
    \frac{1}{{l_\epsilon(g)}}(\sum_{k=1}^{N_{n}} \delta_ \epsilon(g_k,g_i)), & { x_i \in S_n.} 
    \end{array}
    \right.
\end{equation}


{After calculating the gradient norm in a decoupled way, the gradient density harmonizing parameter is defined as

\begin{equation}
\beta_i= \frac{N}{GD(g_i)^{\gamma_i}},
\end{equation}
\begin{equation}
\ {
\gamma_i=\left\{
  \begin{array}{ll}
  {\mu_n,} & {g_i \geq \lambda, x_i \in S_n} \\ 
  {\mu_c,} & {g_i \geq \lambda, x_i \in S_c} \\
  {1,} & \text{otherwise,}
  \end{array}\right.
  }
\end{equation}
where outlier threshold $\lambda \in [0,1]$.

We embed DGHM into CE loss, then DGHM-C loss is formulated as
\begin{equation}
\begin{aligned} L_{D G H M-C} &=\frac{1}{MN} \sum_{i=1}^{N} \beta_{i} L_{C E}\left(p_{i}, p_{i}^{*}\right),
\end{aligned}
\end{equation}
where \textit{M} is the number of gradient norm distributions. It should be noted that modulating factor $\gamma_i \geq 0$ in eq.(11) can control the contribution of outliters in the DGHM-C loss function eq.(12), i.e. $\mu_n \geq 1$ is chosen to down-weight the outliers in $S_n$ to avoid overfitting the noisy examples, and $\mu_c \leq 1$ is considered to up-weight the outliers in $S_c$ to avoid under-learning of clean examples. After applying harmonizing parameter on gradient norm, the reformulated gradient norm of different losses are shown in Figure 4.} It can be seen that the curve of GHM-C loss and two curves of DGHM-C loss, i.e. DGHM-C-clean and DGHM-C-noisy, are of the similar trend before outliers. However, outliers in DGHM-C-clean are relatively up-weighted, which is opposite to GHM-C loss and different from focal loss. On the other hand, outliers in DGHM-C-noisy are explicitly down-weighted, while only slightly down-weighting can be observed in GHM-C loss. 
\begin{figure}
\setlength{\abovecaptionskip}{0.cm}
\setlength{\belowcaptionskip}{-5.cm}
\begin{center}
\includegraphics[width=0.5\textwidth]{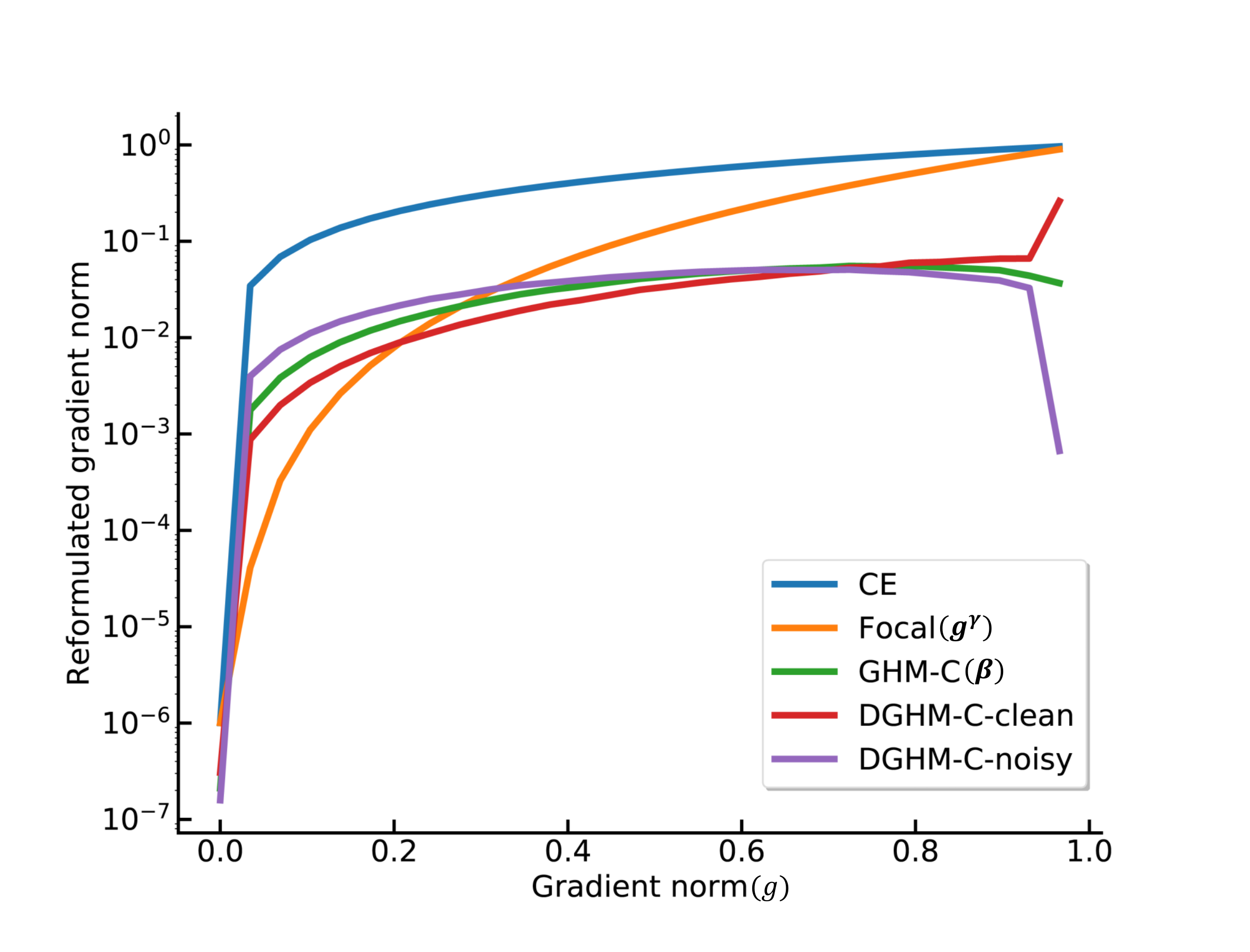}
\caption{Reformulated gradient norm of different loss functions. The y-axis is in log scale. As shown, DGHM-C down-weights the outliers of noisy examples while up-weights those of the clean ones.}
\label{Fig4.}

\end{center}
\end{figure}

\section{Experiment}

\subsection{Dataset}
The dataset is provided by MICCAI Digestpath2019 challenge. A total of 90 patients' 450 pathology images are provided, where 78 AP images with SRC annotation and 372 NP images. The pathology images are from 2 organs, including gastric mucosa and intestine. All pathology images are stained by hematoxylin and eosin (H \& E) and scanned at $\times$40, with the size of 2,000 $\times$ 2,000. Each SRC is annotated by a rectangle bounding box tightly surrounding the cell with a total number of 15,000 cells annotated.
It should be noted that the SRCs are partially annotated, where pathologists can guarantee that the annotated cells are indeed SRCs. The unannotated regions of AP images are also expected to contain SRCs. It is also guaranteed that NP images contains no SRCs. 
 

\subsection{Implement Details}
RetinaNet \cite{focal} with backbone of ImageNet pre-trained ResNet-18 \cite{ResNet} is used in our experiments. We follow the setting in \cite{focal} except the basic areas of anchors, which are tuned for our task from $16^2$ to $256^2$. 
All models are trained with Adam optimizer with initial learning rate of 0.0001, and the learning rate is decreased by 0.1 at 9th and 12th epoch. Since the provided images are too large to feed into the memory, sliding windows strategy is used to decompose the original images into patches of 800 $\times$ 800 pixels with stride of 300 pixels. The patches are then randomly cropped to 600-pixel width and height, followed by horizontal flip and rotation for data augmentation. Models are trained on a single 1080Ti GPU with mini-batch size of 8. To balance the positive and negative samples during training, 1:3 ratio of positive to negative examples is used when constructing a mini-batch.

The whole available images are randomly split into 5 subsets with around 16 positive images and 72 negative images each fold. We perform 5-fold cross validation in the comparison experiment of different loss functions. For other experiments, performance on a randomly chosen subset is evaluated as the result.

\subsection{Evaluation}
Recall and precision are usually used for object detection evaluation \cite{pascal}. In PAOD, however, precision is not reliable because of distinct label noise.
Therefore, same as the challenge, instance-level recall, normal region false positives (NFPs) and FROC \cite{FROC} are used in our experiments with requirement of precision over 0.2. Instance-level recall at IoU of 0.3 is computed among AP images, which is defined as 
\begin{equation}
    Recall = {\frac{TP}{TP + FN}},
\end{equation}
{where \textit{TP} is the number of correctly detected positive examples and \textit{FN} is missed ones. The definition of precision is
\begin{equation}
    Precision = {\frac{TP}{TP + FP}},
\end{equation}
where \textit{FP} is the number of negative examples predicted as positive. However, precision is not reliable, because some examples in \textit{TP} are wrongly attributed as \textit{FP} due to partial annotation.}
Instead, since NP images contain no SRCs, NFPs is considered to measure the wrong predictions of models.
\begin{equation}
    NFPs = \max(100- W,0),
\end{equation}
where \textit{W} is the average number of predicted boxes in NP images.
To access recall and NFPs comprehensively, we further consider FROC \cite{organizer}, which is defined as average recall given confidence threshold at NFPs in S = [1, 2, 4, 8, 16, 32].
\begin{equation}
    FROC =\frac{1}{N}\sum_{i\in S}recall^{NFPs=i}.
\end{equation}

Moreover, we further verify the recall on training data. In controlled label missing rate experiments,
only a part of annotations are used for training and the rest are manually removed. We denote the recall on training and removed annotations as T-recall and R-recall respectively. Intuitively, T-recall reveals whether model has already fit the training data well, while R-recall shows whether unannotated objects can be detected. To some extent, these two metrics can be used to evaluate whether the hard example under-learning and noisy example overfitting are really addressed.

\subsection{Ablation Study}
For a comprehensive understanding of each term in DGHM-C loss, a series of ablation studies are conducted to assess the sensitivity of outlier operations, i.e. the modulating factor $\mu_n$, $\mu_c$ and outlier threshold $\lambda$. With $\mu_n=\mu_c=1$, only decoupling is applied without any outlier operations, which can be considered as the baseline. NFPs are almost unchanged in both 
{\ Table 1 and Table 2}, which is mainly because the classification of NP and AP images is a relatively simple task. Thus, we focus on the recall and FROC, while precision is briefly discussed since the absolute value is meaningless.

\begin{table*}[width=2.0 \linewidth,cols=6,pos=h]
{
\caption{Performance comparison with varying $\mu_n$ and $\mu_c$ ($\lambda=0.9$). The average performance and standard deviation over 5 different fold are provided.}\label{tbl1}
\begin{tabular*}{\tblwidth}{@{} p{0.5cm} LLLLL@{} }
\toprule
$\mu_n$ & $\mu_c$ & Precision & NFPs &  Recall & FROC\\
\midrule
1.0 & 1.0 & 0.4822$\pm$0.0667 & 99.96$\pm$0.0233 & 0.7790$\pm$0.0253 & 0.7716$\pm$0.0265 \\
\hline
2.0 & 1.0 & 0.4756$\pm$0.0633 & 99.98$\pm$0.0310 & 0.7816$\pm$0.0342 &  0.7814$\pm$0.0339 \\
1.0 &  0.5 & 0.4224$\pm$0.0527 & 99.98$\pm$0.0221 &  0.8292$\pm$0.0228  &  0.8134$\pm$0.0274  \\
\hline
0.5 & 2.0 & \textbf{0.4950$\pm$0.0527} & \textbf{99.98$\pm$0.0172} & 0.7590$\pm$0.0258 & 0.7538$\pm$0.0173 \\
1/1.5 & 1.5 & 0.4828$\pm$0.0722 & 99.92$\pm$0.0946 & 0.7742$\pm$0.0371 & 0.7706$\pm$0.0410 \\
1.5 & 1/1.5 & 0.4534$\pm$0.0698 & 99.95$\pm$0.0620 & 0.8004$\pm$0.0300 & 0.7962$\pm$0.0277 \\
2.0 & 0.5 & 0.4229$\pm$0.0658 & 99.92$\pm$0.0459 & \textbf{0.8294$\pm$0.0251}  & \textbf{0.8189$\pm$0.0337} \\
\bottomrule
\end{tabular*}
}
\end{table*}

\begin{table}[width=.9\linewidth,cols=5,pos=h]
\caption{Performance comparison with varying $\lambda$ ({\ $\mu_n=2.0$, $\mu_c=0.5$}). The performance on a random subset is presented.} \label{tbl2}
\begin{tabular*}{\tblwidth}{p{0.5cm} LLLL}
\toprule
$\lambda$ & Precision & NFPs & Recall & FROC\\
\midrule
     $0.7$ & 0.0844 & 99.69 & \textbf{0.9810} & \textbf{0.9796}  \\
     $0.8$ & 0.2888 & \textbf{99.99} & 0.9111  &  0.9045 \\
     $0.9$ & 0.3568 &  99.97 &  0.8732 &  0.8732 \\
     
\bottomrule
\end{tabular*}
\end{table}

{
Firstly, we evaluate the influence of modulating factors $\mu_n$ and $\mu_c$ (varying from 1/2.0 to 2.0) with $\lambda=0.9$ fixed. As shown in Table 1, singly setting $\mu_n=2.0$ or $\mu_c=0.5$ both boost the performance of recall and FROC compared to baseline. It quite makes sense since noisy example overfitting is expected to alleviated by down-weighting ($\mu_n>1.0$) and hard example under-learning is expected to eased by up-weighting ($\mu_c<1.0$). However, the improvement of the former is limited. That is mainly because that down-weighting is already achieved to some extent for this dataset via GD function even without further modulating ($\mu_n=1.0$, $\mu_c=1.0$). We further set $\mu_n>1.0$ and $\mu_c<1.0$ at the same time to relieve both problems, and the performance are further improved as expected. The opposite operation ($\mu_n<1.0$ and $\mu_c>1.0$) definitely deteriorates the performance. The decrease of precision is mainly because more unannotated SRCs are detected. Thus, $\mu_n>1.0$ and $\mu_c<1.0$ is recommended.} 

Secondly, we assess the effect of outlier threshold by varying $\lambda$ from 0.7 to 0.9 with $\mu_n=2.0$, $\mu_c=0.5$ fixed. As shown in Table 2, recall and FROC increase significantly with $\lambda$ decreasing. However, precision also decreases dramatically. When $\lambda=0.7$, precision is even lower than 20\%, which can be considered as a failure result. With low $\lambda$, the model tends to predict more anchors as positive in AP images. In other words, recall and FROC may be improved at the cost of precision. This is because that, extreme penalties are introduced even when the model is not very confident with the results. Intuitively, model stops from learning early, which causes underfitting. To avoid this situation, a relatively high $\lambda$ value is selected. 

We take $\mu_n=2.0$ , $\mu_c=0.5$ and $\lambda=0.9$ as our default settings, which is the same as the challenge submission. It's worth noting that performance could be further improved by tuning hyper parameters. For example, the modulating factors for outliers in $S_c$ and $S_n$ are not necessarily to be reciprocal relationship. However, it's not our main concern. Under the current settings, $\mu_n>1$ improves recall and FROC, and precision is guaranteed by higher $\lambda$, from which the recall-precision trade off can be easily obtained. Other hyper parameters like number of unit region are chosen to be 10 for low computational complexity, which has been discussed in \cite{GHM}.

\subsection{Comparing with Other Losses}

In this experiment, the same network under the same setting is trained with different losses for classification branch. We compare DGHM-C loss with CE loss, focal loss, GHM-C loss and SCE loss, where CE loss is considered as the baseline. 
As can be seen in {\ Table 3}, focal loss deteriorates the performance compared with baseline. This is mainly because noisy examples are mistaken as hard examples, resulting in noisy example overfitting. On the contrary, GHM-C loss improves recall and FROC by nearly 15\%, which largely benefits from down-weighting the outliers. For SCE loss, two hyper parameters $\alpha_{SCE}=0.01$ and $\beta_{SCE}=1.0$ are used to ease overfitting as in \cite{SCE}. Nonetheless, SCE loss brings no benefits compared with baseline even in NFPs. This is likely because that, good performance of SCE is not guaranteed under asymmetric noise in PAOD. 
However, when applying DGHM-C loss, the best performance of NFPs, recall and FROC are achieved, { {where}} The decline of precision is also acceptable as discussed in section 4.4. The substantial improvement of performance is mainly because DGHM-C loss facilitates the adaptive outlier operations for examples in data space of clean($S_c$) and noisy($S_n$). The problem of hard example under-learning is alleviated by up-weighting the outliers in $S_c$ to encourage learning, while the problem of noisy example overfitting is addressed by down-weighting the outliers in $S_n$ to enhance noise tolerance. This can also be proved by qualitative result in Figure 5. Furthermore, we have also tried applying different losses to noisy and clean examples respectively, such as combination of focal/CE loss with GHM-C loss. It is found that tuning weighting factor of different losses is quite trivial since their magnitude are distinct. The convergence of model is not guaranteed. In contrast, models trained with DGHM-C loss also yield better convergence.

\begin{table*}[width=2.0 \linewidth,cols=5,pos=h]
\caption{Quantitative results of RetinaNet with different loss functions. The average performance and standard deviation of 5 different fold is presented.}
\label{tbl3}
\begin{tabular*}{\tblwidth}{@{} p{2cm} LLLL@{} }
\toprule
Method & Precision & NFPs & Recall & FROC\\
\midrule
CE & 0.6088$\pm$0.0741 &  99.37$\pm$0.2059 & 0.5960$\pm$0.0258 & 0.5931$\pm$0.0251  \\
Focal & 0.6556$\pm$0.0791 & 99.26$\pm$0.4685 & 0.5086$\pm$0.0197 & 0.5027$\pm$0.0214 \\
GHM-C & 0.4968$\pm$0.0678 & 99.72$\pm$0.1867  & 0.7565$\pm$0.0396 & 0.7429$\pm$0.0395  \\
SCE & 0.5926$\pm$0.0703 & 97.90$\pm$0.5922 & 0.5779$\pm$0.0191 & 0.5594$\pm$0.0178   \\
DGHM-C & 0.4229$\pm$0.0658 & \textbf{99.92}$\pm$0.0459   & \textbf{0.8294}$\pm$0.0251 & \textbf{0.8188}$\pm$0.0337  \\

\bottomrule
\end{tabular*}
\end{table*}

\begin{figure*}
\setlength{\abovecaptionskip}{-0.cm}
\setlength{\belowcaptionskip}{-0.cm}
\begin{center}
\includegraphics[width=0.95\textwidth]{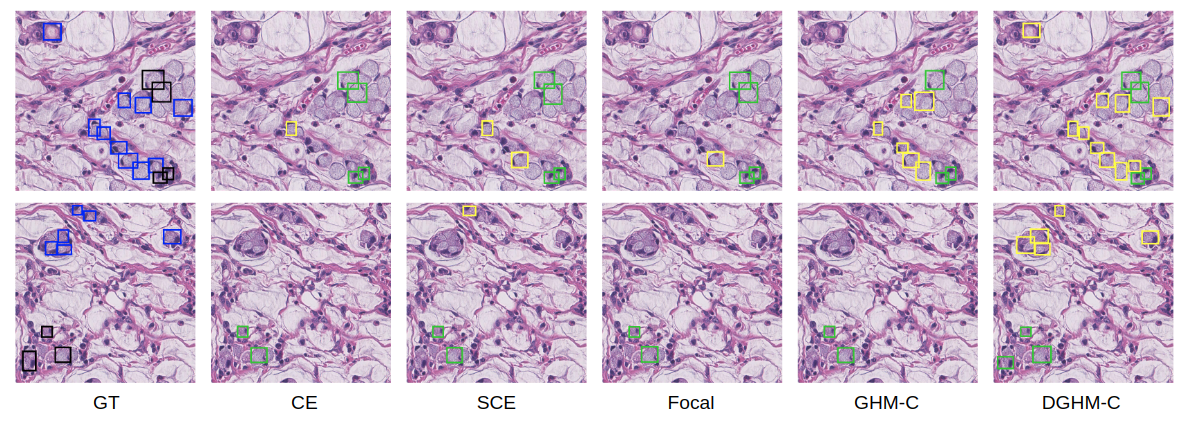}
\caption{
Qualitative comparison of our method and four references(GHM-C loss, focal loss, CE loss and SCE loss) under 70\% missing label rates. The first column shows the ground truth, where black bounding boxes are training annotations and the blue ones are removed annotations. From second column, green bounding boxes indicate annotated SRC correctly detected by models, while the yellow ones correspond to unannotated SRC detected.
}
\label{Fig5.}
\end{center}
\end{figure*}

\subsection{Controlled Label Missing Rate} \label{sec:clmr}

To investigate the effect of missing label rate $\eta$, we randomly discard the provided annotation by rate of $[20\%, 70\%]$. For simplicity, we assume the provided annotation is perfect and take $\eta$ as the annotation drop rate. The performance of different methods is listed in {\ Table 4}.
As can be seen, our method achieves outstanding recall from 0.6917 to 0.8666 and FROC from 0.6720 to 0.8630 under different $\eta$, which outperforms other methods by a significant margin. Furthermore, when changing $\eta$ from 20\% to 70\%, for model trained with focal loss, CE loss and SCE loss, the FROC performance significantly drop by nearly 35\%. Though GHM-C loss shows a better tolerance to noise, it still suffers a 25\% drop in FROC performance. However, a 20\% drop is obtained by DGHM-C loss, which is a much smaller gap compared with others. The same behavior is also observed for recall.

As shown in {\ Table 4}, R-recall of our DGHM-C loss surpasses other methods by over 15\% under varying $\eta$. In the extreme case of dropping 70\% of annotations, our method registers nearly 30\% improvement (0.6872 vs. 0.3872) compared with the closest competitor (GHM-C), from which we can conclude that our model is more robust to label noise even when $\eta$ is large. As for T-recall, though it is a relatively simple task, 3\% to 10\% enhancement are achieved by DGHM-C loss, which indicates our model can learn hard examples well.
When considering R-recall and T-recall together, one can find CE loss, focal loss and SCE loss are all with relatively high T-recall but low R-recall. (The R-recall of 0.6590 for SCE under $\eta=40\%$ is considered as an outlier.) This is mainly because that they suffer from noisy example overfitting. Though relative high R-recall is acquired by GHM-C loss, the T-recall is unstable, e.g. T-recall is 0.6446 when $\eta=70\%$. It seems that the noise robustness is achieved at cost of hard example under-learning. Only through DGHM-C loss, both good performance of T-recall and R-recall can be accomplished.

Moreover, we present a few qualitative results in Figure 5 for models trained with $\eta=70\%$. 
As can be seen, the left bottom SRC in figures of the second row is missed by other methods, indicating it might be a hard example, while our method successfully detect it. Moreover, our method captures more removed SRCs (yellow bounding boxes), illustrating the robustness of our model to partial annotation.



\begin{table*}[width=2.14\linewidth,cols=4,pos=h]
\caption{Performance comparison under different missing label rates on a single subset. Recall, precision, FROC and NFPs metrics are evaluated on full-annotated test set.}\label{tbl4}

\begin{tabular*}{\tblwidth}{l|cccccc|cccccc}
\toprule
\hline
\multicolumn{1}{l|}{\multirow{2}{*}{Method}} & \multicolumn{6}{c|}{\textbf{Precision}} & \multicolumn{6}{c}{\textbf{NFPs}} \\ \cline{2-13}
  & 20\%                   & 30\%                   & 40\%                   & 50\%                   & 60\%                   & 70\% & 20\% & 30\%                   & 40\%                   & 50\%                   & 60\%                   & 70\%                   \\ \hline
\multicolumn{1}{l|}{CE}     & \multicolumn{1}{l}{0.5803}    & \multicolumn{1}{l}{0.6096} & \multicolumn{1}{l}{0.6020} & \multicolumn{1}{l}{0.6211} & \multicolumn{1}{l}{0.6445} & \multicolumn{1}{l|}{0.6117} &
  \multicolumn{1}{l}{99.54}    & \multicolumn{1}{l}{99.77} & \multicolumn{1}{l}{99.73} & \multicolumn{1}{l}{99.59} & \multicolumn{1}{l}{99.81} & \multicolumn{1}{l}{99.67} \\  
\hline
\multicolumn{1}{l|}{Focal} & \multicolumn{1}{l}{0.6077}    & \multicolumn{1}{l}{0.6167} & \multicolumn{1}{l}{0.6359} & \multicolumn{1}{l}{0.6074} & \multicolumn{1}{l}{0.6552} & \multicolumn{1}{l|}{0.5878} &
\multicolumn{1}{l}{99.53}  & \multicolumn{1}{l}{99.64} & \multicolumn{1}{l}{99.63} & \multicolumn{1}{l}{99.46} & \multicolumn{1}{l}{99.62} & \multicolumn{1}{l}{99.67} \\ \hline  
 \multicolumn{1}{l|}{GHM-C} & \multicolumn{1}{l}{0.4795}    & \multicolumn{1}{l}{0.4992} & \multicolumn{1}{l}{0.5539} & \multicolumn{1}{l}{0.4949} & \multicolumn{1}{l}{0.6267}  & \multicolumn{1}{l|}{0.6867}
& \multicolumn{1}{l}{99.91}    & \multicolumn{1}{l}{99.97} & \multicolumn{1}{l}{\textbf{100.0}} & \multicolumn{1}{l}{99.92} & \multicolumn{1}{l}{100.0}  & \multicolumn{1}{l}{\textbf{100.0}} \\
\hline

\multicolumn{1}{l|}{SCE}  & \multicolumn{1}{l}{0.5428}    & \multicolumn{1}{l}{0.5835} & \multicolumn{1}{l}{0.5818} & \multicolumn{1}{l}{0.5893} & \multicolumn{1}{l}{0.5933}  &
\multicolumn{1}{l|}{0.6006} &
\multicolumn{1}{l}{97.97}    & \multicolumn{1}{l}{98.95} & \multicolumn{1}{l}{98.51} & \multicolumn{1}{l}{98.67} & \multicolumn{1}{l}{99.45} & \multicolumn{1}{l}{99.47} \\
\hline
\multicolumn{1}{l|}{DGHM-C} & 
\multicolumn{1}{l}{0.3785}    & \multicolumn{1}{l}{0.3155}  & \multicolumn{1}{l}{0.3702} & \multicolumn{1}{l}{0.4071}  & \multicolumn{1}{l}{0.4609} & \multicolumn{1}{l|}{0.4229} &
\multicolumn{1}{l}{\textbf{99.91}}    & \multicolumn{1}{l}{\textbf{99.98}} & \multicolumn{1}{l}{99.99} & \multicolumn{1}{l}{\textbf{99.96}} & \multicolumn{1}{l}{\textbf{100.0}} & \multicolumn{1}{l}{99.96} \\ \hline

\hline
 & \multicolumn{6}{c|}{\textbf{Recall}} & \multicolumn{6}{c}{\textbf{FROC}} \\ \hline

\multicolumn{1}{l|}{CE}     & 
\multicolumn{1}{l}{0.5532} & \multicolumn{1}{l}{0.4519} & \multicolumn{1}{l}{0.387} & \multicolumn{1}{l}{0.3513} & \multicolumn{1}{l}{0.2682} & \multicolumn{1}{l|}{0.1902} &
\multicolumn{1}{l}{0.5255} & \multicolumn{1}{l}{0.4490} & \multicolumn{1}{l}{0.3870} & \multicolumn{1}{l}{0.3513} & \multicolumn{1}{l}{0.2682} & \multicolumn{1}{l}{0.1866} \\ \hline
\multicolumn{1}{l|}{Focal}  & 
\multicolumn{1}{l}{0.4585} & \multicolumn{1}{l}{0.3870} & \multicolumn{1}{l}{0.2762} & \multicolumn{1}{l}{0.2289} & \multicolumn{1}{l}{0.1939} & \multicolumn{1}{l|}{0.1050} &
\multicolumn{1}{l}{0.4541} & \multicolumn{1}{l}{0.3856} & \multicolumn{1}{l}{0.2719} &
\multicolumn{1}{l}{0.2172} & \multicolumn{1}{l}{0.1837} & \multicolumn{1}{l}{0.1050}  \\ \hline
\multicolumn{1}{l|}{GHM-C}  &
\multicolumn{1}{l}{0.6642} & \multicolumn{1}{l}{0.6983} & \multicolumn{1}{l}{0.6480} & \multicolumn{1}{l}{0.6727} & \multicolumn{1}{l}{0.5138} &\multicolumn{1}{l|}{0.4089} &  

\multicolumn{1}{l}{0.6545} & \multicolumn{1}{l}{0.6968} & \multicolumn{1}{l}{0.6480} & \multicolumn{1}{l}{0.6669} & \multicolumn{1}{l}{0.5138} & \multicolumn{1}{l}{0.4089} \\ \hline
\multicolumn{1}{l|}{SCE}  & 
\multicolumn{1}{l}{0.5175} & \multicolumn{1}{l}{0.4278} & \multicolumn{1}{l}{0.4249} & \multicolumn{1}{l}{0.3440} & \multicolumn{1}{l}{0.2340} & \multicolumn{1}{l|}{0.1545} &

\multicolumn{1}{l}{0.4928} & \multicolumn{1}{l}{0.4253} & \multicolumn{1}{l}{0.4147} & \multicolumn{1}{l}{0.3352} & \multicolumn{1}{l}{0.2340} & \multicolumn{1}{l}{0.1505}  \\ \hline

\multicolumn{1}{l|}{DGHM-C} & 
\multicolumn{1}{l}{\textbf{0.8666}} & \multicolumn{1}{l}{\textbf{0.8476}} & \multicolumn{1}{l}{\textbf{0.8105}} & \multicolumn{1}{l}{\textbf{0.7682}} & \multicolumn{1}{l}{\textbf{0.7128}} & \multicolumn{1}{l|}{\textbf{0.6917}} &

\multicolumn{1}{l}{\textbf{0.8630}} & \multicolumn{1}{l}{\textbf{0.8331}} & \multicolumn{1}{l}{\textbf{0.7922}} & \multicolumn{1}{l}{\textbf{0.7412}} & \multicolumn{1}{l}{\textbf{0.7128}} & \multicolumn{1}{l}{\textbf{0.6720}} \\ \hline


\hline
 & \multicolumn{6}{c|}{\textbf{R-recall}} & \multicolumn{6}{c}{\textbf{T-recall}} \\ \hline

\multicolumn{1}{l|}{CE}     & \multicolumn{1}{l}{0.3612} & \multicolumn{1}{l}{0.2353} & \multicolumn{1}{l}{0.2131} & \multicolumn{1}{l}{0.1845} & \multicolumn{1}{l}{0.1477} & \multicolumn{1}{l|}{0.1036}     & \multicolumn{1}{l}{0.9105}    & \multicolumn{1}{l}{0.9033} & \multicolumn{1}{l}{0.9078} & \multicolumn{1}{l}{0.9070} & \multicolumn{1}{l}{0.8989} & \multicolumn{1}{l}{0.9054} \\ \hline
\multicolumn{1}{l|}{Focal}  & \multicolumn{1}{l}{0.2268} & \multicolumn{1}{l}{0.1431} & \multicolumn{1}{l}{0.1029} & \multicolumn{1}{l}{0.0925} & \multicolumn{1}{l}{0.0597} & \multicolumn{1}{l|}{0.0397}    & \multicolumn{1}{l}{0.8932}    & \multicolumn{1}{l}{0.8843} & \multicolumn{1}{l}{0.8821} & \multicolumn{1}{l}{0.8865} & \multicolumn{1}{l}{0.8844} & \multicolumn{1}{l}{0.8496} \\ \hline
 \multicolumn{1}{l|}{GHM-C}  & \multicolumn{1}{l}{0.7556} & \multicolumn{1}{l}{0.7044} & \multicolumn{1}{l}{0.6547} & \multicolumn{1}{l}{0.6754} & \multicolumn{1}{l}{0.5233} & \multicolumn{1}{l|}{0.3872}     & \multicolumn{1}{l}{0.8341}    & \multicolumn{1}{l}{0.7525} & \multicolumn{1}{l}{0.8616} & \multicolumn{1}{l}{0.7194} & \multicolumn{1}{l}{0.8509}  & \multicolumn{1}{l}{0.6446} \\ \hline
\multicolumn{1}{l|}{SCE} & \multicolumn{1}{l}{0.4607} & \multicolumn{1}{l}{0.3725} & \multicolumn{1}{l}{0.6590} & \multicolumn{1}{l}{0.2595} & \multicolumn{1}{l}{0.2002} & \multicolumn{1}{l|}{0.1117}     & \multicolumn{1}{l}{0.8960}    & \multicolumn{1}{l}{0.8816} & \multicolumn{1}{l}{0.8899} & \multicolumn{1}{l}{0.8851} & \multicolumn{1}{l}{0.8736} & \multicolumn{1}{l}{0.8544} \\ \hline

\multicolumn{1}{l|}{DGHM-C} & \multicolumn{1}{l}{\textbf{0.8560}} & \multicolumn{1}{l}{\textbf{0.8693}} & \multicolumn{1}{l}{\textbf{0.8067}} & \multicolumn{1}{l}{\textbf{0.7761}} & \multicolumn{1}{l}{\textbf{0.6836}} & \multicolumn{1}{l|}{\textbf{0.6872}}    & \multicolumn{1}{l}{\textbf{0.9340}}    & \multicolumn{1}{l}{\textbf{0.9903}} & \multicolumn{1}{l}{\textbf{0.9935}} & \multicolumn{1}{l}{\textbf{0.9887}} & \multicolumn{1}{l}{\textbf{0.9876}} & \multicolumn{1}{l}{\textbf{0.9900}} \\ \hline

\end{tabular*}
\end{table*}




\section{Discussion}

\subsection{Decoupling More}

Though great performance is achieved by decoupling noisy examples from clean ones, we believe that the optimal decoupling method is hard to define. As an additional research, we further decouple clean positive examples $\textit{\text{AP}}_P$ and clean negative examples {$\textit{\text{NP}}_N$}, applying similar DGHM with {\ $\mu_n=2.0$} for both $\textit{\text{AP}}_P$ and { $\textit{\text{NP}}_N$}. We denote it as DGHM-C$^*$. 
The comparison result is shown in {\ Table 5}. We observe that DGHM-C$^*$ achieves a 3\% improvement on recall and FROC compared to DGHM-C. To analyze the potential reason for the performance boost, we illustrate the according gradient norm distributions in Figure 6. As can be seen,
a large parts of examples from { $\textit{\text{NP}}_N$} show a very low gradient norm, which is consistent with the high NFPs in above experiments. It indicates that learning to distinguish NP images from abnormal ones is relatively easy for model. If the gradient density is shared between positive and negative examples, their loss contribution are the same. However, if the gradient density is calculated separately, the loss contribution of examples from $\textit{\text{AP}}_P$ are larger than that from { $\textit{\text{NP}}_N$}, even the same whole gradient norm is unchanged. The model is forced to learn to correctly classify positive from negative.
From empirical understanding, more delicate decoupling methods potentially boost DGHM, resulting in better performance. Still, computational complexity should be taken into consideration.

\begin{table*}[width=2.1\linewidth,cols=5,pos=h]
\caption{Performance difference between DGHM-C and DGHM-C$^*$. The average performance and standard deviation of 5 different fold are reported. For both methods, we set hypermeters {\ $\mu_n=2$, $\mu_c=0.5$}  and $\lambda=0.9$. }
\label{tbl1}
\begin{tabular*}{\tblwidth}{@{} LLLLL@{} }
\toprule
Method & Precision & NFPs & Recall & FROC\\
\midrule
DGHM-C & 0.4229$\pm$0.0658 & 99.92$\pm$0.0455   & 0.8294$\pm$0.0252 & 0.8188$\pm$0.0338 \\
DGHM-C$^*$ & 0.3722$\pm$0.0683 & \textbf{99.96$\pm$0.0249} & \textbf{0.8533$\pm$0.0366} & \textbf{0.8494$\pm$0.0372}  \\
\bottomrule
\end{tabular*}
\end{table*}

\begin{figure}
\setlength{\abovecaptionskip}{-0.cm}
\setlength{\belowcaptionskip}{-0.cm}
\begin{center}
\includegraphics[width=0.45\textwidth]{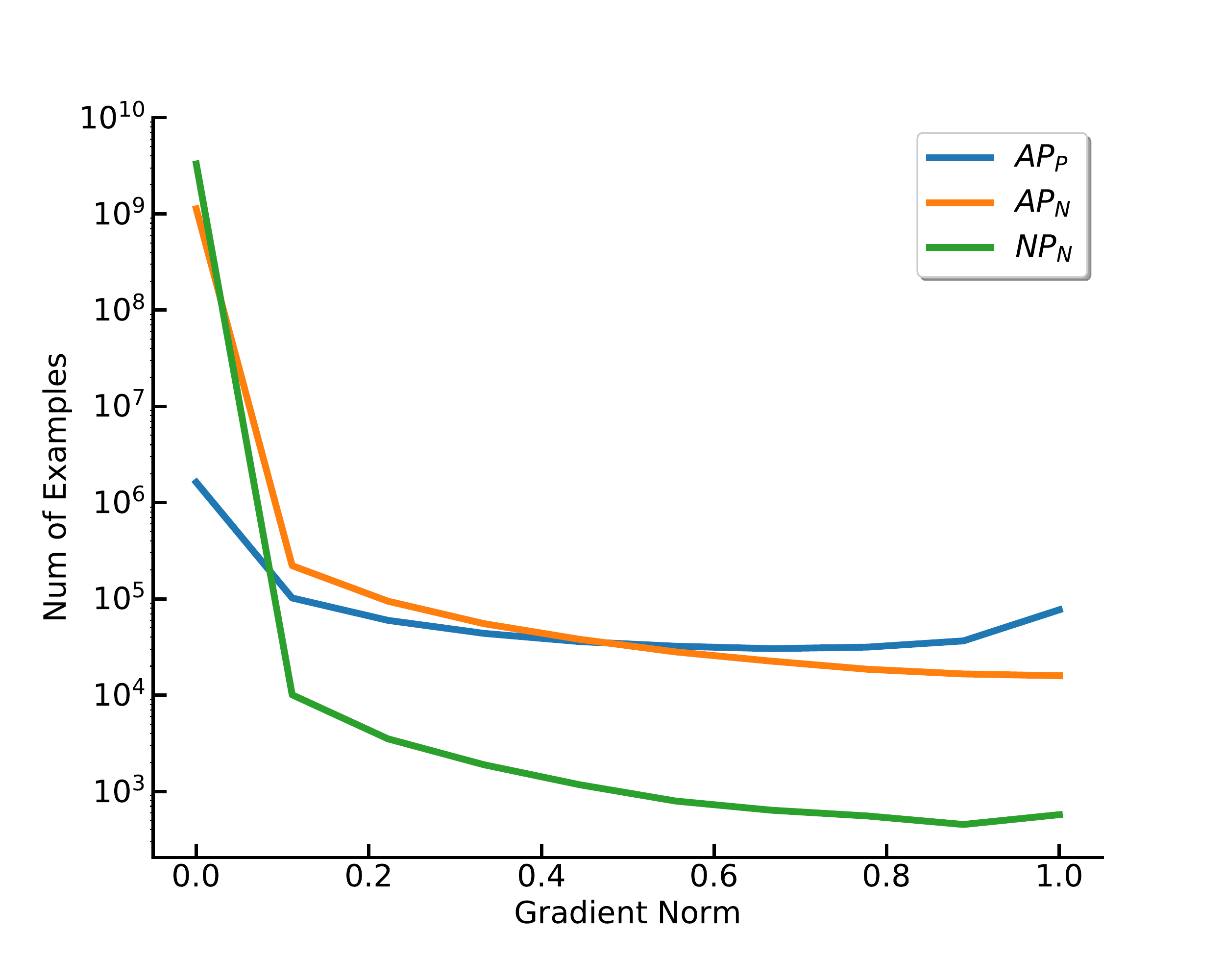}
\caption{
Gradient norm distribution of a converged model on $\textit{\text{AP}}_P$, $\textit{\text{AP}}_N$ and { $\textit{\text{NP}}_N$}. The y-axis is in log scale.
}
\label{Fig6.}
\end{center}
\end{figure}

\subsection{Recall/Precision Trade-off}
A trade-off between recall and precision can be observed in DGHM-C loss.
Actually, we can manually settle the recall/precision trade-off in the way of tuning the hyper parameters of the modulating factor {\ $\mu_n$, $\mu_c$} and outliers threshold $\lambda$.
However, a model is hard to simultaneously achieves both high recall and good precision in above experiments. 
The main reason as mentioned before is that SRCs are partially annotated, some correctly detected SRCs are counted as false positives. Therefore, the measured precision is not accurate, which is lower than the real one.
Besides, bounding boxes regression could be another reason.
Though unannotated SRCs may be detected in classification by DGHM-C loss, their regression are never trained, which leads to localization failure during non-maximum suppression. It will be our future work to improve recall and precision at the same time.

\section{Conclusion}

In this paper, we formulate the partial annotation problem as an integrated task of noisy-supervised learning and full-supervised learning. To solve the problem, we propose a novel Decoupled Gradient Harmonizing Mechanism (DGHM) and embed it into classification loss (DGHM-C). Thus, both hard example under-learning and noisy example overfitting are addressed simultaneously. 
Experiments show that our method surpasses other baselines on MICCAI DigesPath2019 challenge. We further demonstrate the efficacy of our method by reporting an extensive controlled missing rates experiment. In the future we would like to explore the decoupling mechanism and regression branch of PAOD.
\bibliographystyle{cas-model2-names}

\bibliography{main}

\newpage

\end{document}